\documentclass[10pt,twocolumn,letterpaper]{article}

\usepackage{cvpr}              % To produce the CAMERA-READY version
\usepackage{graphicx}
\usepackage{amsmath}
\usepackage{amssymb}
\usepackage{booktabs}

\usepackage[pagebackref,breaklinks,colorlinks]{hyperref}

\usepackage[capitalize]{cleveref}
\crefname{section}{Sec.}{Secs.}
\Crefname{section}{Section}{Sections}
\Crefname{table}{Table}{Tables}
\crefname{table}{Tab.}{Tabs.}

\usepackage{amssymb}% http://ctan.org/pkg/amssymb
\usepackage{pifont}% http://ctan.org/pkg/pifont
\newcommand{\checkmarkx}{\ding{55}}%
\usepackage{xcolor}
\newcommand{\firstbest}[1]{\textcolor{red}{\textbf{#1}}}
\newcommand{\secondbest}[1]{\textcolor{blue}{#1}}
\newcommand{\colorlegend}{\bigskip \raggedright \footnotesize Legend: \firstbest{best}, \secondbest{second best} result.}%

\def\cvprPaperID{20} % *** Enter the CVPR Paper ID here
\def\confName{CVPRW: NTIRE} 
\def\confYear{2024}

\begin{document}

\title{Unsupervised Image Prior via Prompt Learning and CLIP Semantic Guidance for Low-Light Image Enhancement}

\author{Igor Morawski$^1$ \quad  Kai He$^3$  \quad Shusil Dangi$^3$  \quad  Winston H. Hsu$^{1,2}$  
\\
\\
$^1$National Taiwan University \qquad $^2$Mobile Drive Technology \qquad $^3$Qualcomm Inc. 
}

\maketitle

%%%%%%%%% ABSTRACT
\begin{abstract}
   Currently, low-light conditions present a significant challenge for machine cognition. In this paper, rather than optimizing models by assuming that human and machine cognition are correlated, we use zero-reference low-light enhancement to improve the performance of downstream task models. We propose to improve the zero-reference low-light enhancement method by leveraging the rich visual-linguistic CLIP prior without any need for paired or unpaired normal-light data, which is laborious and difficult to collect. We propose a simple but effective strategy to learn prompts that help guide the enhancement method and experimentally show that the prompts learned without any need for normal-light data improve image contrast, reduce over-enhancement, and reduce noise over-amplification. Next, we propose to reuse the CLIP model for semantic guidance via zero-shot open vocabulary classification to optimize low-light enhancement for task-based performance rather than human visual perception. We conduct extensive experimental results showing that the proposed method leads to consistent improvements across various datasets regarding task-based performance and compare our method against state-of-the-art methods, showing favorable results across various low-light datasets.
\end{abstract}

\begin{figure}
    \centering
    \small
    \begin{subfigure}{.15\textwidth}
      \centering
        \includegraphics[width=1\linewidth]{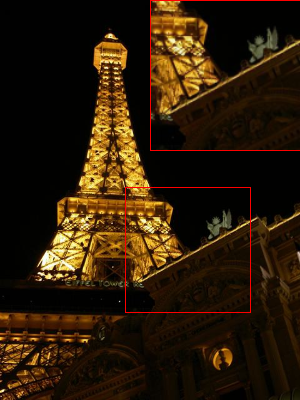}
        \caption*{}
    \end{subfigure} %
    \begin{subfigure}{.15\textwidth}
      \centering
        \includegraphics[width=1\linewidth]{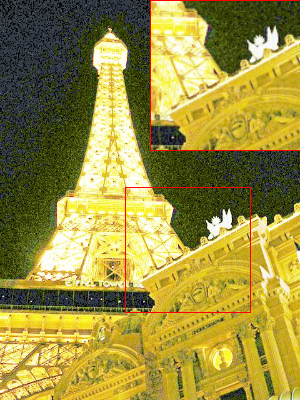}
        \caption*{}
    \end{subfigure} %
    \begin{subfigure}{.15\textwidth}
      \centering
        \includegraphics[width=1\linewidth]{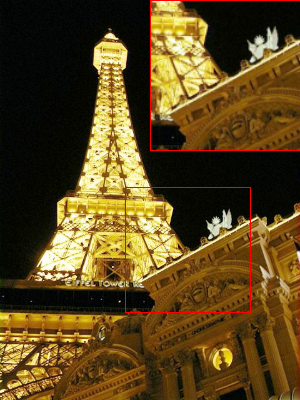}
        \caption*{}
    \end{subfigure} %

\vspace{-1.2em}

    \begin{subfigure}{.15\textwidth}
      \centering
        \includegraphics[width=1\linewidth]{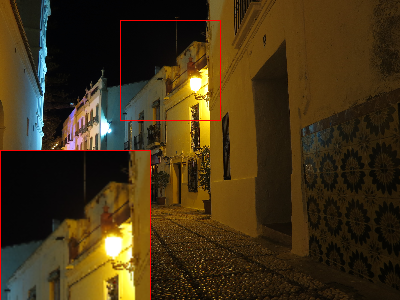}
        \caption*{ Input}
    \end{subfigure} %
    \begin{subfigure}{.15\textwidth}
      \centering
        \includegraphics[width=1\linewidth]{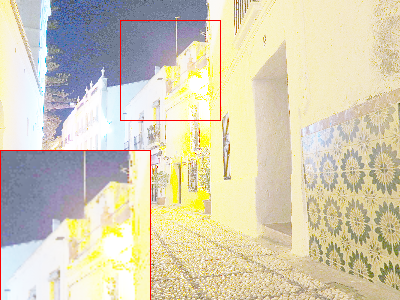}
        \caption*{ Baseline \cite{guo2020zero}}
    \end{subfigure} %
    \begin{subfigure}{.15\textwidth}
      \centering
        \includegraphics[width=1\linewidth]{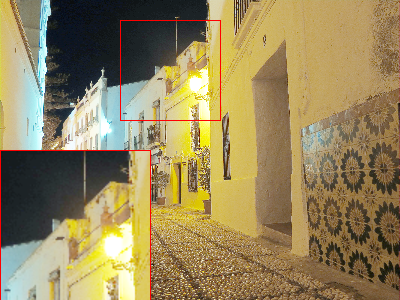}
        \caption*{ Ours}
    \end{subfigure} %

\vspace{-1em}

    \caption{Our proposed method leverages the CLIP \cite{radford2021learning} model for unsupervised image prior via prompt learning and open-vocabulary semantic guidance. Our proposed method improves the over-all image hue, reduces over-enhancement and reduces noise over-amplification. Further, we conduct extensive experiments to show that our proposed method significantly improves machine cognition as measured by task-based performance of down-stream tasks models, without incurring any additional computation costs on the light-weight enhancement baseline model \cite{guo2020zero}.}
    \label{fig:first-gifure}
\end{figure}

\begin{figure*}[]
    \centering
    \small
    \includegraphics[width=1\linewidth]{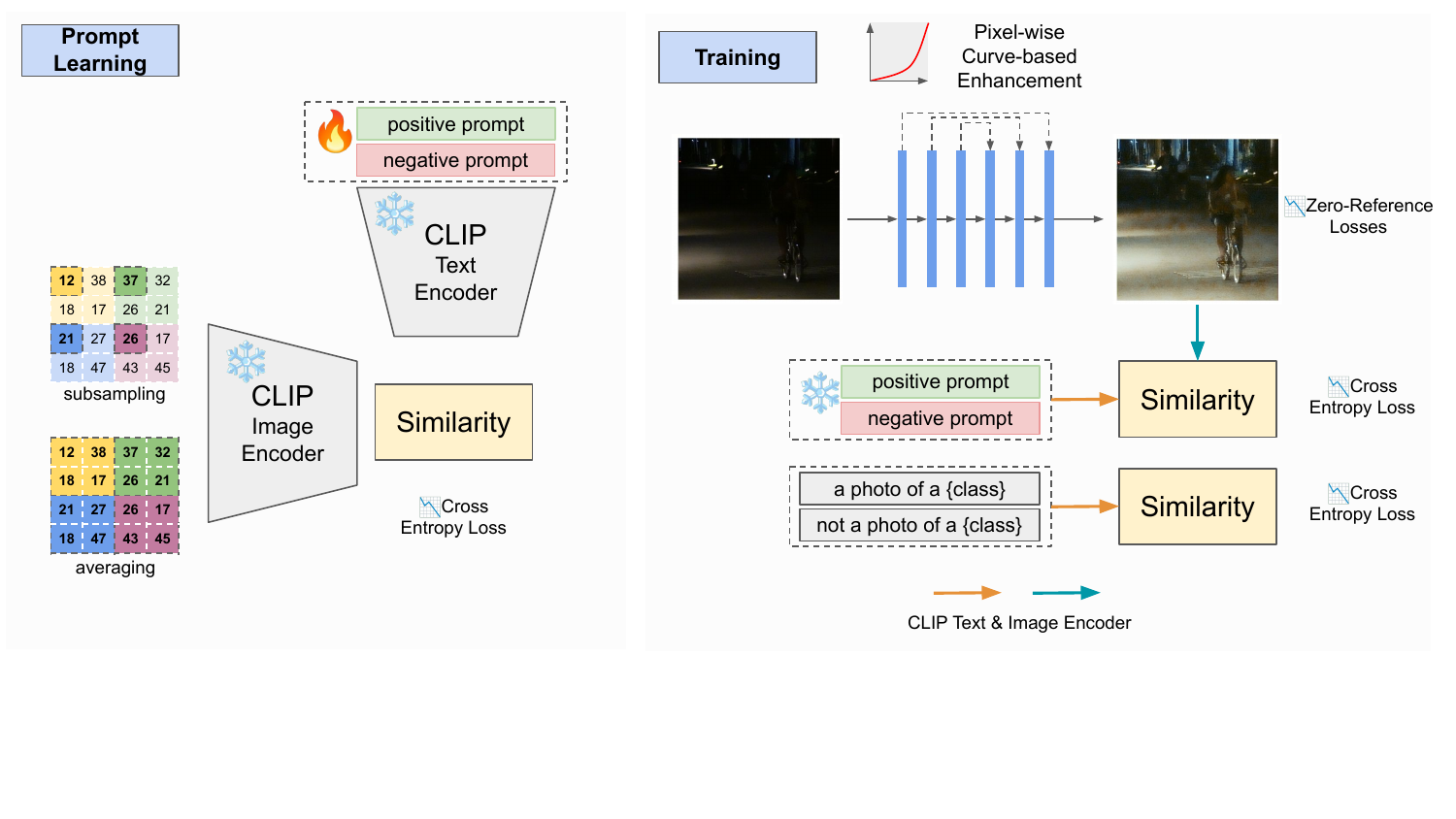}
    \vspace{-2.5cm}
    \caption{Our proposed two-stage training process leverages the pre-trained CLIP model that can capture lighting conditions and quality of images. We propose to use the CLIP model to learn the positive and negative image priors with a simple data augmentation strategy without any need for paired or unpaired normal-light data via prompt learning, and use them for guiding the image enhancement model. During the training, we use the learned prompts and reuse the CLIP model for semantic guidance to improve the quality of the enhanced images. Our proposed method uses open-vocabulary classification, so it can be easily extended to any dataset, without limiting object categories, with annotated bounding boxes or any type of annotation that can be used to extract patches with an object category, as well as to paired low- and normal-light datasets, increasing the variety of the data in the training. }
    \label{fig:proposed-method}
\end{figure*}

%-------------------------------------------------------------------------
\section{Introduction}
\label{sec:introduction}
Low-light conditions present a significant challenge for image quality, adversely impacting the performance of high-level computer vision models. In addition to high-level noise caused by photon-limited imaging, low-light images often contain out-of-focus and motion blur, unnatural appearance caused by camera flash and image signal processor failures such as incorrect white-balancing or tone-mapping, all affecting the image quality negatively. With the increased integration of high-level computer vision methods into daily life,  addressing the robustness of machine cognition in low-light conditions becomes ever more important. While most works try to improve machine cognition by correlating the human visual perception with the image quality, we study machine cognition-oriented low-light image enhancement in this paper.

Many works \cite{xu2020learning,zheng2021adaptive,fan2022half,SNRAware,Wu_2023_CVPR,Xu_2023_CVPR,wei2018deep,zhang2019kindling,zhang2021beyond,yang2021sparse,zhang2022deep} propose to supervise the training using paired low- and normal-light data. However, the collection of paired low- and normal-light data is laborious because it usually requires capturing the same scene with different sensor exposure time. Because of the collection costs of low- and normal-light data, some works \cite{yang2020fidelity,jiang2021enlightengan,liang2023iterative,guo2020zero,li2021learning} propose to learn image enhancement in an unsupervised fashion. While some methods \cite{yang2020fidelity,jiang2021enlightengan,liang2023iterative} still require the selection of unpaired low- and normal-light datasets, zero-reference methods such as \cite{guo2020zero,li2021learning,zheng2022semantic} only require low-light data for training. Motivated by difficulties in collecting paired normal- and low-light data, we focus on zero-reference image enhancement in this work. While zero-reference methods  \cite{guo2020zero,li2021learning,zheng2022semantic} solve the problems associated with collection costs, they do not integrate semantic knowledge during the training process.

Motivated by the zero-shot capabilities of open vocabulary image understanding \cite{zang2022open,kuo2022f,zhou2022extract,cheng2024yolo} enabled by the CLIP \cite{radford2021learning} model, we incorporate the rich CLIP prior that can capture the lighting conditions and image quality \cite{liang2023iterative,wang2023exploring} into the zero-reference training process. 

Our contributions are as follows: 
\begin{itemize}
    \item We propose to learn a low-light image prior via prompt learning, using a simple data augmentation strategy, without any need for paired or unpaired normal-light data. The proposed prompts help to guide the image enhancement model by improving the contrast, reducing loss of information caused by over-enhancement of bright regions, and reducing over-amplification of noise.
    \item We reuse the CLIP model for semantic guidance via open vocabulary image classification. While our strategy is simple, it scales favorably to include datasets with any object categories. In addition to that, paired low- and normal-light datasets can also be included in training by using high-scoring detections inferred by a  generic object detector on normal-light data.
    \item We conduct extensive ablation and comparison experiments to show that our proposed method leads to consistent improvements in task-based performance across many high-level low-light datasets.
\end{itemize}

%-------------------------------------------------------------------------

\section{Related Work}
\label{sec:related_work}

\subsection{Low-Light Enhancement}
\textbf{Traditional methods}. Traditional approaches to low-light enhancement include histogram-based and Retinex-based methods. Histogram-based methods employ a mapping of intensity values based on histograms at a global \cite{gonzalez2009digital,wang2007fast} or local \cite{kim2001advanced,stark2000adaptive} level. While simple and efficient, traditional histogram-based methods typically discount structural information, do not address the problem of denoising and are prone to producing unnatural artifacts including color shift. On the other hand, methods based on the Retinex theory \cite{land1977retinex} aim to decompose a low-light image into reflectance and illumination components and treat the estimated reflectance or its modification as the final output. Seminal works single-scale \cite{jobson1997properties} and multi-scale \cite{jobson1997multiscale} Retinex propose to estimate image illumination with single- and multi-scale Gaussian filters. Over the years, more complex Retinex-based methods such as \cite{wang2013naturalness,fu2016fusion,guo2016lime} have been proposed, including joint enhancement and denoising methods \cite{li2017joint,li2018structure}, to produce more natural-looking results. Because traditional Retinex-based methods rely on hand-crafted features and assumptions about images, they require careful parameter tuning, might not generalize well to the variety of real-life images, and often lack robustness against image degradation.

\textbf{Learning-based methods}. In recent years, motivated by impressive results demonstrated by deep learning methods in computer vision, many deep learning low-light image enhancement methods have been proposed. Existing methods can be generally divided into end-to-end methods \cite{xu2020learning,jiang2021enlightengan,zheng2021adaptive,fan2022half,dong2022abandoning,SNRAware,Xu_2023_CVPR,Wu_2023_CVPR} that directly enhance the input image and Retinex-based methods  \cite{wei2018deep,zhang2019kindling,zhang2021beyond,liu2021retinex,yang2021sparse,wang2022low,wu2022uretinex,ma2022toward,fu2023you,liang2023iterative,zhang2022deep} that typically decompose the image into reflectance and illumination components. Many of these methods require large amounts of paired low- and normal-light images to train \cite{xu2020learning,zheng2021adaptive,fan2022half,SNRAware,Wu_2023_CVPR,Xu_2023_CVPR,wei2018deep,zhang2019kindling,zhang2021beyond,yang2021sparse,zhang2022deep}. While synthetic data is easy to generate, models trained on it have difficulties generalizing to real-life images. On the other hand, real data requires significant collection and annotation efforts because the paired data is collected either by post-processing images by trained experts \cite{bychkovsky2011learning} or varying camera and lighting settings \cite{chen2018learning,wei2018deep,cai2018learning,chen2019seeing,hai2023r2rnet}. Although it is possible to use a beam splitter to capture the same scene with two different sensors \cite{jiang2019learning,ExLPose_2023_CVPR}, the majority of the existing datasets are collected by capturing the same scene multiple times with varying exposure settings, and are thus limited to static scenes. Motivated by the costs and limitations of real paired datasets, some works \cite{yang2020fidelity,jiang2021enlightengan,liang2023iterative} propose to use unpaired low- and normal-light data instead. However, careful selection of normal-light data is still necessary. Consequently, Guo \textit{et al.} \cite{guo2020zero,li2021learning} proposed a zero-reference method which does not require any paired or unpaired normal-light data. In \cite{guo2020zero,li2021learning,zheng2022semantic} Guo \textit{et al.} formulated image enhancement as a curve estimation problem and proposed a set of zero-reference losses based on assumptions about natural images, such as average image brightness and gray world hypothesis. 

\subsection{Low-Light Image Understanding}
Loh and Chan \cite{loh2019getting} have experimentally demonstrated that convolutional neural networks trained on low- and normal-light data for image recognition do not learn to normalize deep image features. That is, Loh and Chan \cite{loh2019getting} showed that extracted low- and normal-light features belong to distinct feature clusters. Since then, low-light image understanding has further received the interest of researchers in high-level tasks such as face detection \cite{poor_visibility_benchmark,liang2021recurrent,wang2021hla}, object detection \cite{morawski2021nod,morawski2022genisp,hong2021crafting,loh2019getting}, pose estimation \cite{ExLPose_2023_CVPR} and action recognition \cite{xu2021arid}. % Some works focus on improving performance of task-specific models by improving model robustness \cite{ExLPose_2023_CVPR,li2021photon} or domain adaptation \cite{Kennerley_2023_CVPR,wang2021hla,sasagawa2020yolo}.
Another line of works focuses on improving machine cognition by introducing low-light enhancement to the cognition framework and learning the enhancement under supervision of a high-level task model \cite{liang2021recurrent,morawski2021nod,wang2023tienet,zheng2022semantic,hashmi2023featenhancer}. In contrast with previous works in image enhancement and restoration that use semantic information for guidance at a loss\cite{liu2017image,aakerberg2022semantic} or feature level \cite{wang2018recovering,li2020blind}, task-oriented enhancement models \cite{liang2021recurrent,morawski2021nod,wang2023tienet,zheng2022semantic,hashmi2023featenhancer} optimize for machine perception rather than assuming correlation between human perception and downstream task performances \cite{morawski2021nod,morawski2022genisp,wang2023tienet,hashmi2023featenhancer,robidoux2021end,yoshimura2023dynamicisp}.  Moreover, as an enhanced image can be reused by many different generic downstream-task models, this approach might be attractive in resource-limited scenarios. 

\subsection{CLIP in Vision}
Following the remarkable success of Contrastive Language-Image Pre-Training (CLIP) \cite{radford2021learning}, the CLIP visual-linguistic prior has been leveraged to generalize to open-vocabulary problems in high-level vision \cite{zang2022open,kuo2022f,zhou2022extract,cheng2024yolo}. Recently, Wang \textit{et al.} \cite{wang2023exploring} showed that CLIP can be used to capture the quality and abstract \textit{feel} of the images, and Liang \textit{et al.} \cite{liang2023iterative} showed that the CLIP prior can differentiate between various lighting conditions in natural images. Motivated by this, Liang \textit{et al.} \cite{liang2023iterative} proposed to leverage the CLIP prior for image enhancement by iteratively learning prompts to discriminate between normal-light and back-lit images. 

As we focus on low-light machine cognition, we make minimal assumptions about correlating the human and machine visual perception and employ task-based low-light performance to evaluate our proposed enhancement method.

\begin{figure}[h]
    \centering
    \small
    \begin{subfigure}{.45\textwidth}
      \centering
        \includegraphics[width=1\linewidth]{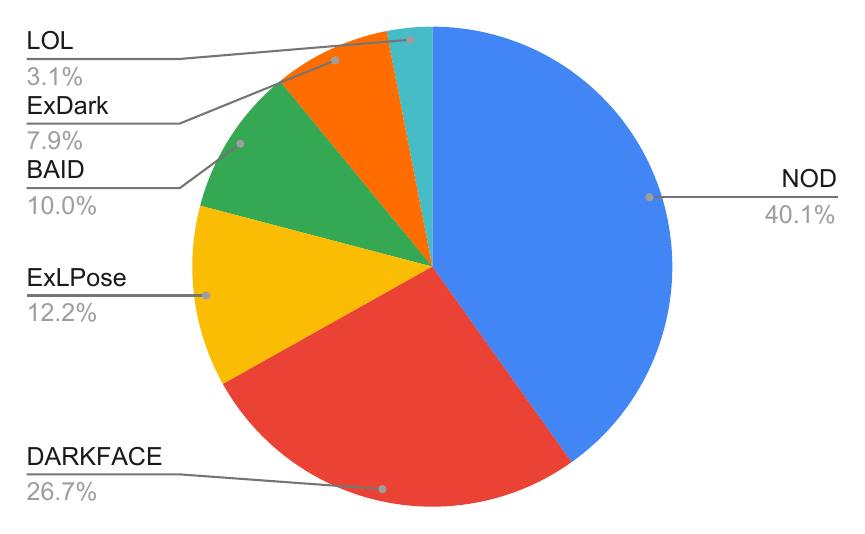}
        \caption{Proportions of samples in the dataset.}
    \end{subfigure}%
    
    \begin{subfigure}{.45\textwidth}
      \centering
        \includegraphics[width=1\linewidth]{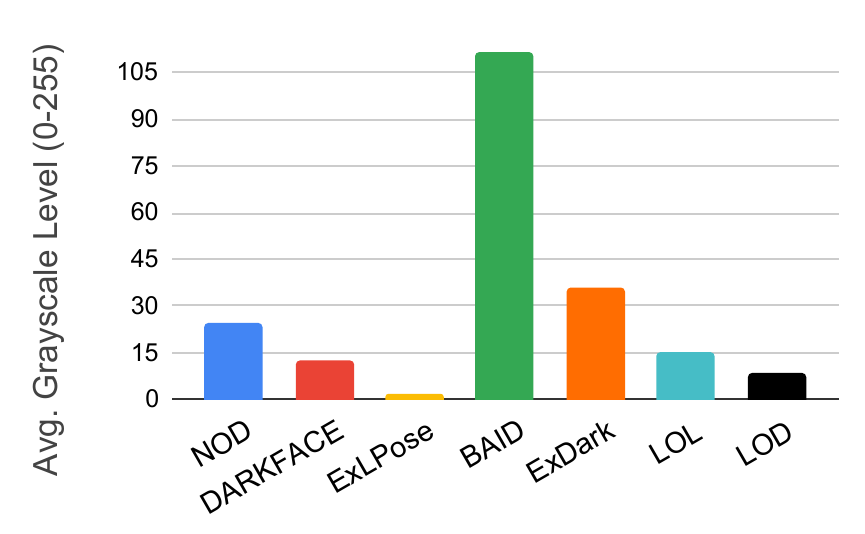}
        \caption{Average image brightness statistics.}
    \end{subfigure}%
    \caption{Statistics of the datasets used for the ablation study.}
    \label{fig:training-stats}
\end{figure}

%-------------------------------------------------------------------------

\section{Proposed Method}
\label{sec:proposed_method}
We propose a two-stage training process that leverages the pre-trained CLIP model for semantic guidance and learning an image prior from low-light data without any additional paired or unpaired normal-light data. 

During the pre-training stage, we propose a simple but effective strategy to learn the pair of positive and negative image prompts to help to guide the enhancement. Next, we use the learned prompts and additionally reuse the CLIP model for semantic guidance to improve the enhancement of low-light images without any need for paired or unpaired low-light data. Because our proposed semantic guidance is realized by open-vocabulary CLIP classification, the proposed method can be easily extended to any low-light dataset containing bounding-box annotations or any type of annotation that can be used to extract patches with an object category.

\subsection{Unsupervised Image Prior via Prompt Learning}
\label{ssec:learnable-image-prior}
\begin{figure}[]
    \centering
    \small
    \includegraphics[width=0.4\linewidth]{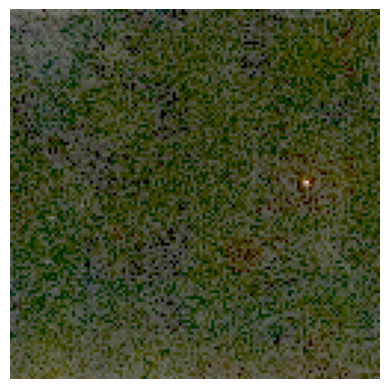}
    \includegraphics[width=0.4\linewidth]{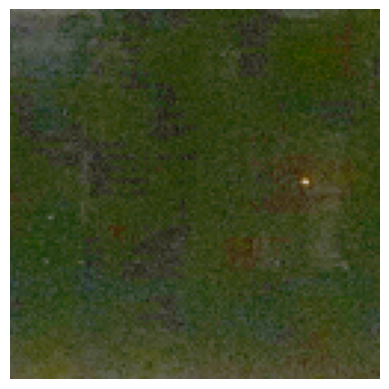}
    % \vspace{-2.5cm}
    \caption{Our strategy to learn positive and negative image prompts:  $1:4$ subsampled (left) negative and $4 \times 4$ averaged (right) positive image. Subsampling preserves the noise in the image while averaging acts as a fast and simply proxy for denoising. }
    \label{fig:sampling}
\end{figure}

Motivated by the observation that the CLIP image prior can capture diverse lighting conditions and image quality, we propose to leverage CLIP to learn an image prior from low-light data without the need for paired or unpaired normal-light data. During the first stage of training, we use the pre-trained CLIP model to learn positive and negative prompts to discriminate between averaged and subsampled low-light data, respectively. 

Given a low-light image $I \in \mathbb{R}^{H\times W \times C}$ and a pair of randomly initialized positive and negative prompts, $P_p \in \mathbb{R}^{N \times 512}$ and $P_n \in \mathbb{R}^{N \times 512}$, where $N$ denotes the prompt length, we first augment the image using random photometric augmentation into $I'$ to avoid learning prompts that would  overly constraint the overall image brightness. Next, we generate a pair of positive and negative images. 

We generate a positive image $I_p = avg_{m \times m}(I')$ by applying $m \times m$ average pooling to the augmented image $I'$ as a fast and simple proxy for denoising. We generate a negative image $I_n = sub_{1:m}(I')$ by applying $1:m$ subsampling to the augmented image $I'$, keeping the original noise levels. The process is illustrated in Fig. \ref{fig:proposed-method} and the effect of averaging and subsampling is showed in Fig. \ref{fig:sampling}.

We use the binary cross-entropy loss to learn the prompt pair to capture the image prior while differentiating between the image quality.

\begin{equation}
  L_{prompt\ init.} = -( y \log{ \hat{y} } + (1-y) \log{ (1-\hat{y}) } ),
  \label{eq:prompt-learning-ce}
\end{equation}

where $y$ is the image label (0 for a positive, $4 \times 4$ averaged and 1 for a negative, $1:4$ subsampled image) and $\hat{y}$ is based on the softmax of cosine similarity between each of the prompts and each of the images $I \in \{I_p, I_n\}$:

\begin{equation}
  \hat{y} = \frac
  {e^{cos(\Phi_{img}(I), \Phi_{txt}(P_p)}}
  {\sum_{i \in \{p, n\}} 
 e^{cos(\Phi_{img}(I), \Phi_{txt}(P_i)}},
  \label{eq:prompt-learning-cs}
\end{equation}

where $\Phi_{img}$ is the CLIP image encoder and $\Phi_{txt}$ is the CLIP text encoder. 

\subsection{Unsupervised Low-Light Enhancement}
Motivated by the success of curve-based enhancement \cite{guo2020zero} for zero-reference low-light enhancement, we adopt DCE-Net \cite{guo2020zero} as our lightweight baseline model that formulates enhancement as an pixel-wise curve prediction applied to the input image $I$ as below:

\begin{equation}
  LE_n(x) = LE_{n-1}(x) + A_n (x) LE_{n-1}(x)(1-LE_{n-1}(x)),
  \label{eq:iterative-curve}
\end{equation}
where $A$ is a set of $N$ pixel-wise parameter maps $\alpha \in [-1, 1]$ applied iteratively to the input image $I$.

\subsubsection{Zero-Reference Image Losses}
\label{sssec:zero-reference-losses}
We follow Guo \textit{et al.} \cite{guo2020zero} and adopt a set of zero-reference loss functions for low-light image enhancement. For clarity, we briefly discuss the employed exposure control $L_{exp}$, spatial consistency $L_{spa}$, color constancy $L_{RGB}$ and illumination smoothness $L_{TV_{A}}$ below.

\textbf{Exposure control loss} $L_{exp}$ encourages exposure correction of under- and over-exposed regions by setting an expected average region intensity $E$:

\begin{equation}
  L_{exp} = \frac{1}{M}
  \sum^{M}_{i=1} 
  \sum_{i=1} 
  |\hat{I}_i - E|,
  \label{eq:loss-exposure}
\end{equation}

where  $\hat{I}_{k}$ is the $k$-th $16 \times 16$ patch of the enhanced image, $M$ is the number of non-overlapping $16 \times 16$ image patches and $E$ is the expected average intensity. $E$ is empirically set to $0.6$

\textbf{Spatial consistency loss} $L_{spa}$ ensures spatial coherence of the enhanced image by preserving relative differences between adjacent image regions:

\begin{equation}
  L_{spa} = \frac{1}{K}
  \sum^{K}_{i=1} 
  \sum_{j \in \Omega (i)} 
  ( |\hat{I}_{i} - \hat{I}_{j}| -
  |{I}_{i} - {I}_{j}| )^2,
  \label{eq:loss-spatial}
\end{equation}

where  ${I}$ and $\hat{I}$ are the original and enhanced images, respectively, $K$ is the number of local regions, and $\Omega (i)$ denotes the 4-connected neighborhood centered at the $4 \times 4$ region $i$.

\textbf{Color loss} $L_{RGB}$ follows the Gray-World hypothesis \cite{buchsbaum1980spatial} to constraint differences between color channels:

\begin{equation}
  L_{RGB} = (\hat{I}_{R,\mu} - \hat{I}_{G,\mu})^2 + (\hat{I}_{R,\mu} - \hat{I}_{B,\mu})^2 + (\hat{I}_{G,\mu} - \hat{I}_{B,\mu})^2
  \label{eq:loss-color}
\end{equation}

where $\hat{I}_{c,\mu}$ is the average intensity of color channel $c \in \{R,G,B\}$ of the enhanced image $\hat{I}$.

\textbf{Illumination smoothness loss} $L_{TV_{A}}$ encourages monotonicty relation between neighboring image pixels and is applied to intermediate curve parameter map $A$:

\begin{equation}
  L_{TV_{A}} =
  \sum_{c\in\{R,G,B\}}
  (|\nabla_x A_c| + |\nabla_y A_c|)^2,
  \label{eq:loss-tv}
\end{equation}

where $A_{c}$ is the intermediate channel-wise curve parameter used to enhance the image $\hat{I}$, and $\nabla$ denotes the gradient operation.

\textbf{Additional zero-reference losses}. We have experimented with additional zero-reference losses, such as total-variation smoothness loss applied to the enhanced image $\hat{I}$, total-variation smoothness loss applied to $\mu$-tonemapped enhanced image $\hat{I}$, comparing differently sampled enhanced image $\hat{I}$ and input image $\hat{I}$. While the additional constraints we tried led to improved performance for some datasets, we found that the impact is not consistent across many datasets and sometimes lead to significant decrease in performance.

\begin{figure*}[t]
    \centering
    \small
    \begin{subfigure}{.19\textwidth}
      \centering
        \includegraphics[width=1\linewidth]{Figures/ablation-figures/CROP_000000_ORIG_DICM_01.png}
        \caption*{}
    \end{subfigure} %
    \begin{subfigure}{.19\textwidth}
      \centering
        \includegraphics[width=1\linewidth]{Figures/ablation-figures/CROP_240302_011148_CLIP0_ZDCE_CROP_ENSEMBLE_batch8x224_ENSEMBLE_ZDCE_BASELINE_DICM_01.png}
        \caption*{}
    \end{subfigure} %
    \begin{subfigure}{.19\textwidth}
      \centering
        \includegraphics[width=1\linewidth]{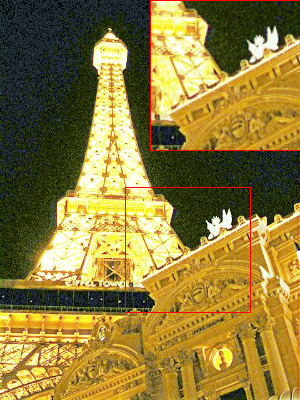}
        \caption*{}
    \end{subfigure} %
    \begin{subfigure}{.19\textwidth}
      \centering
        \includegraphics[width=1\linewidth]{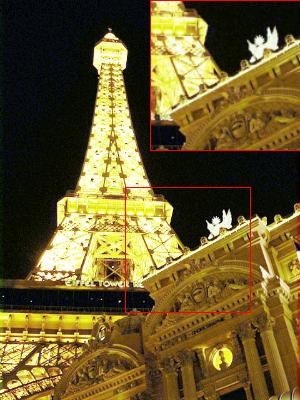}
        \caption*{}
    \end{subfigure} %
    \begin{subfigure}{.19\textwidth}
      \centering
        \includegraphics[width=1\linewidth]{Figures/ablation-figures/CROP_240302_230648_CLIP1_ZDCE_CROP_ENSEMBLE_batch8x224_ENSEMBLE_ZDCE_OPEN_VOCAB_PROMPT_BRIGHTNESS_AUGMENTED12_DICM_01.png}
        \caption*{}
    \end{subfigure} %

\vspace{-1.2em}

    \begin{subfigure}{.19\textwidth}
      \centering
        \includegraphics[width=1\linewidth]{Figures/ablation-figures/CROP_000000_ORIG_LIME_5.png}
    \end{subfigure} %
    \begin{subfigure}{.19\textwidth}
      \centering
        \includegraphics[width=1\linewidth]{Figures/ablation-figures/CROP_240302_011148_CLIP0_ZDCE_CROP_ENSEMBLE_batch8x224_ENSEMBLE_ZDCE_BASELINE_LIME_5.png}
    \end{subfigure} %
    \begin{subfigure}{.19\textwidth}
      \centering
        \includegraphics[width=1\linewidth]{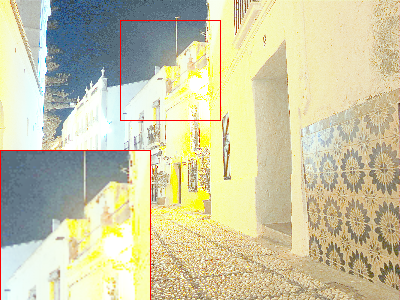}
    \end{subfigure} %
    \begin{subfigure}{.19\textwidth}
      \centering
        \includegraphics[width=1\linewidth]{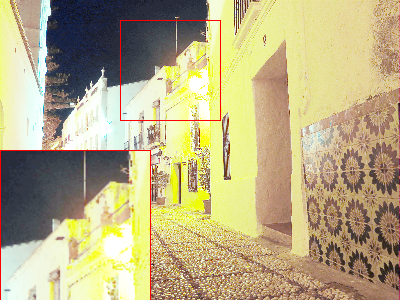}
    \end{subfigure} %
    \begin{subfigure}{.19\textwidth}
      \centering
        \includegraphics[width=1\linewidth]{Figures/ablation-figures/CROP_240302_230648_CLIP1_ZDCE_CROP_ENSEMBLE_batch8x224_ENSEMBLE_ZDCE_OPEN_VOCAB_PROMPT_BRIGHTNESS_AUGMENTED12_LIME_5.png}
    \end{subfigure} %

\vspace{0.2em}

    \begin{subfigure}{.19\textwidth}
      \centering
        \includegraphics[width=1\linewidth]{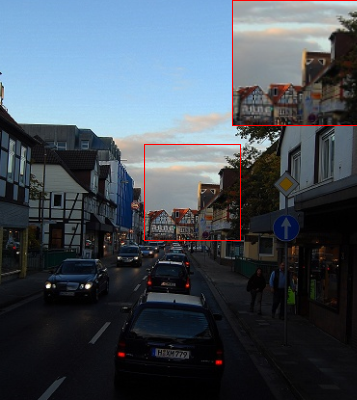}
        \caption*{Input}
    \end{subfigure} %
    \begin{subfigure}{.19\textwidth}
      \centering
        \includegraphics[width=1\linewidth]{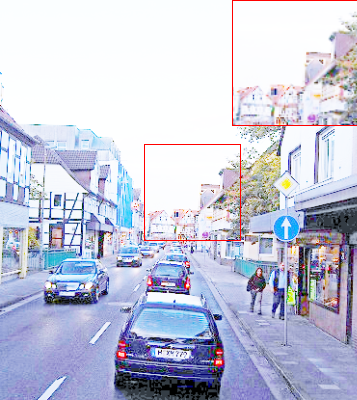}
        \caption*{Baseline \cite{guo2020zero}}
    \end{subfigure} %
    \begin{subfigure}{.19\textwidth}
      \centering
        \includegraphics[width=1\linewidth]{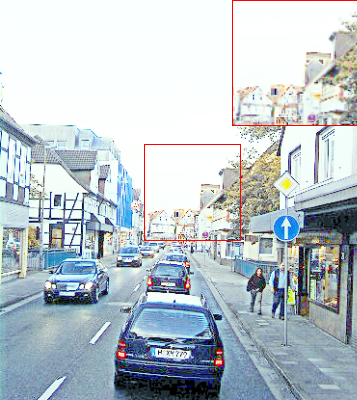}
        \caption*{+ semantic guidance}
    \end{subfigure} %
    \begin{subfigure}{.19\textwidth}
      \centering
        \includegraphics[width=1\linewidth]{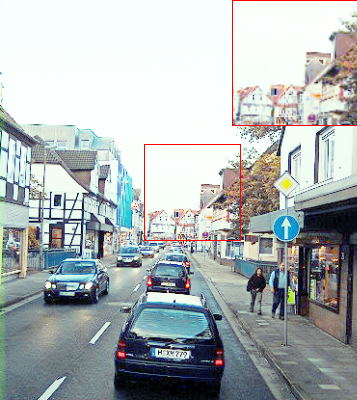}
        \caption*{+ learned prompt}
    \end{subfigure} %
    \begin{subfigure}{.19\textwidth}
      \centering
        \includegraphics[width=1\linewidth]{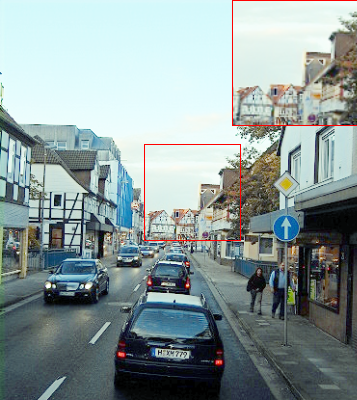}
        \caption*{ Ours}
    \end{subfigure} %

\vspace{-1em}

    \caption{Ablation study of our proposed method. Semantic guidance improves the color distribution of the images, while the learned prompt improves image contrast, reduces overexposure and reduces over-amplification of the noise.}
    \label{fig:ablation-qualitative}
\end{figure*}
% &  &  & 42.1\% & 37.8\% & \firstbest{40.3\%} & 22.8\% & 33.3\% \\

\begin{table*}[t]
\centering
\footnotesize
\begin{tabular}{ccc|cccccc}
\hline

Zero-ref. & Open vocab. & Learned & NOD \cite{morawski2021nod} & NOD SE \cite{morawski2021nod} &  LOD \cite{hong2021crafting} & ExDark \cite{loh2019getting} & ExDark \cite{loh2019getting} & DarkFace \cite{poor_visibility_benchmark} \\

baseline \cite{guo2020zero} & class. & img. prior & mAP $\uparrow$ & mAP $\uparrow$ & mAP $\uparrow$ & mAP $\uparrow$ & class. acc. $\uparrow$ & mAP@.5 $\uparrow$ \\ \hline

\multicolumn{3}{c|}{(no enhancement)} & 42.1\% & 22.8\% & 37.8\% & \firstbest{40.3\%} & 22.8\% & 33.3\%  \\

\checkmark &  &  & 40.4\% & 22.0\% & 38.6\% & 33.3\% & 40.9\% & 33.8\%  \\

\checkmark & \checkmark &  & 41.1\% & 22.5\% & 42.5\% & 34.5\% & 41.2\% & 28.6\% \\

\checkmark &  & \checkmark & \secondbest{42.4\%} & \secondbest{23.3\%} & \secondbest{47.1\%} & 38.0\% & \secondbest{42.7\%} & \secondbest{35.5\%}  \\

\checkmark & \checkmark & \checkmark & \firstbest{43.2\%} & \firstbest{26.5\%} & \firstbest{47.4\%} & \secondbest{39.7\%} & \firstbest{43.6\%} & \firstbest{38.0\%}  \\ 

\hline
\end{tabular}
\\ \colorlegend
\caption{Quantitative results of the ablation study of our proposed method in terms of task-based performance. Our proposed improvements leads to consistent improvement over the baseline method. }
\label{tab:ablation}
\end{table*}

\subsubsection{Leveraging CLIP for Image Enhancement}
\label{sssec:leveraging-clip-training}
We propose to leverage the pre-trained CLIP model which can capture lighting conditions, image quality and semantic information for image enhancement at the loss level.

\textbf{Learned CLIP Image Priors}. To further constrain the enhanced image $\hat{I}$ and improve the quality of the enhanced images, we use the learned image prompts introduced in subsection \ref{ssec:learnable-image-prior}. We found that the image prior learned via prompt learning using our methods reduces over-amplification of the image noise and overexposure of bright regions. 

To this end, we first encode the enhanced image $\hat{I}$ using the CLIP image encoder $\Phi_{img}$ and the learned prompt pair using the CLIP text encoder $\Phi_{txt}$. Next, we compute the cosine similarity between the embedded enhanced image $\Phi_{img}(\hat{I})$ and the learned prompt pair $\Phi_{txt}(P)$, where $P={P_p,P_n}$ is a pair of the positive and negative learned prompts. 

\begin{equation}
  \hat{y}_{prompt} = \frac
  {e^{cos(\Phi_{img}(\hat{I}), \Phi_{txt}(P_p)}}
  {\sum_{i \in \{p, n\}} 
 e^{cos(\Phi_{img}(\hat{I}). \Phi_{txt}(P_i)}},
  \label{eq:prompt-image-cs}
\end{equation}

Finally, we compute the cross-entropy loss, assuming that the enhanced image should match the positive prompt $P_p$.

\begin{figure*}[t]
    \centering
    \small
    \begin{subfigure}{.12\textwidth}
      \centering
        \includegraphics[width=1\linewidth]{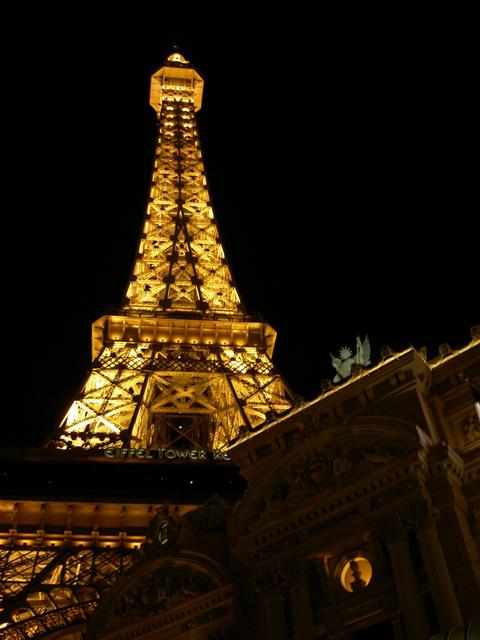}
        \caption*{}
    \end{subfigure} %
    \begin{subfigure}{.12\textwidth}
      \centering
        \includegraphics[width=1\linewidth]{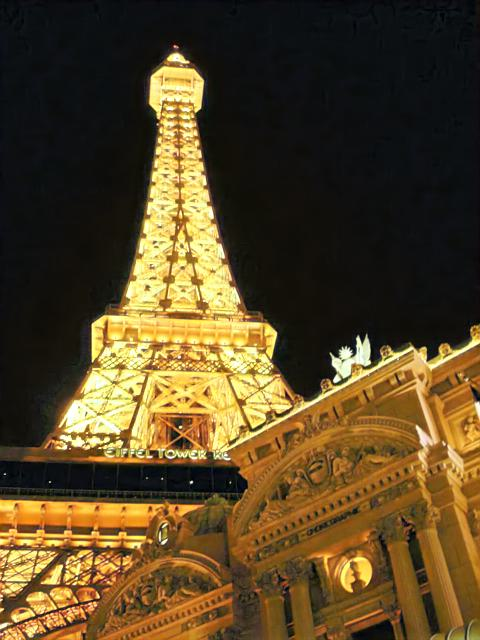}
        \caption*{}
    \end{subfigure} %
    \begin{subfigure}{.12\textwidth}
      \centering
        \includegraphics[width=1\linewidth]{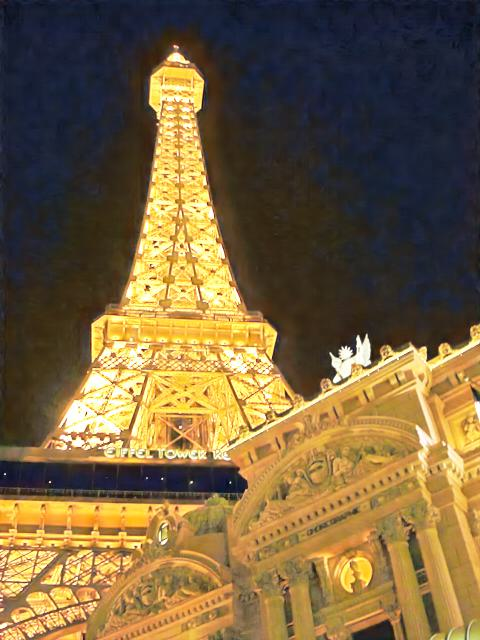}
        \caption*{}
    \end{subfigure} %
    \begin{subfigure}{.12\textwidth}
      \centering
        \includegraphics[width=1\linewidth]{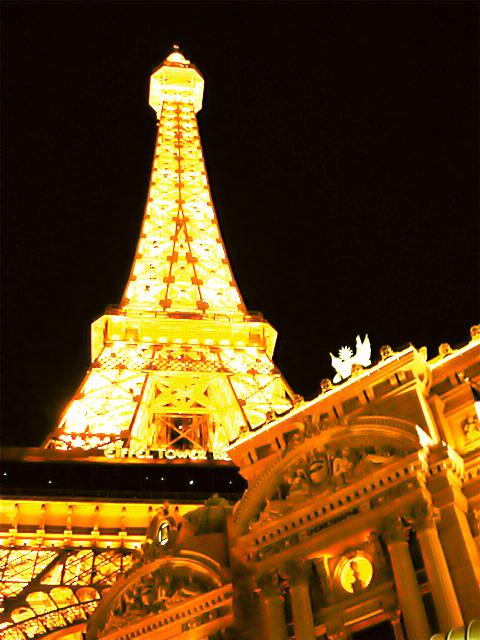}
        \caption*{}
    \end{subfigure} %
    \begin{subfigure}{.12\textwidth}
      \centering
        \includegraphics[width=1\linewidth]{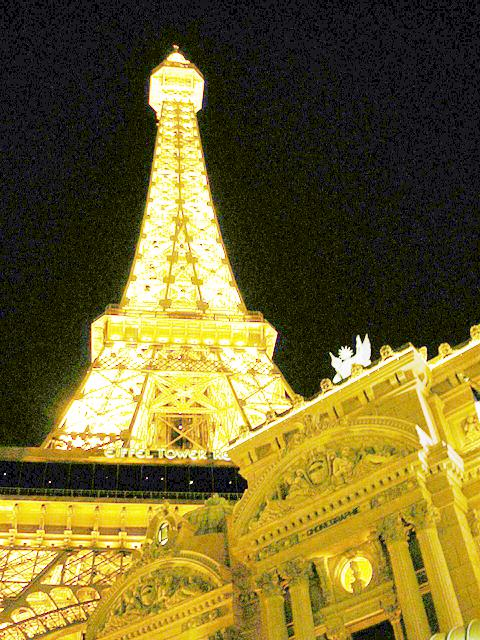}
        \caption*{}
    \end{subfigure} %
    \begin{subfigure}{.12\textwidth}
      \centering
        \includegraphics[width=1\linewidth]{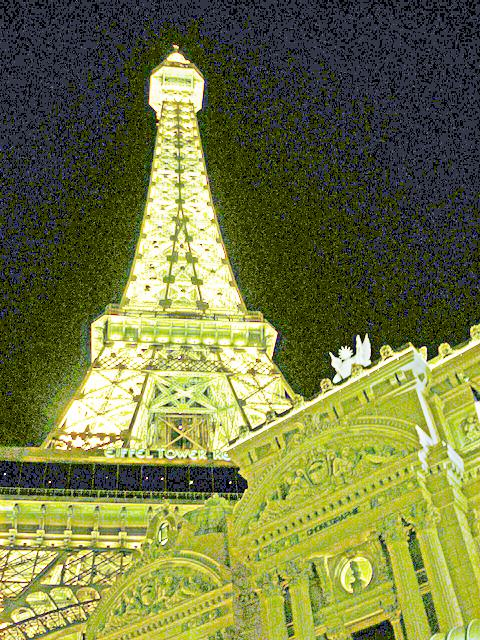}
        \caption*{}
    \end{subfigure} %
    \begin{subfigure}{.12\textwidth}
      \centering
        \includegraphics[width=1\linewidth]{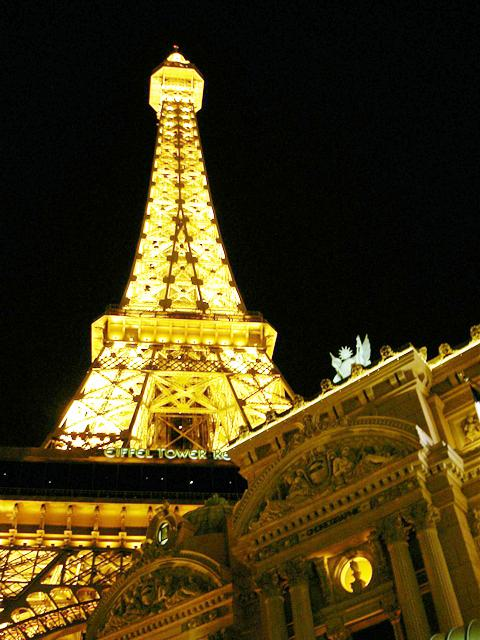}
        \caption*{}
    \end{subfigure} %
    \begin{subfigure}{.12\textwidth}
      \centering
        \includegraphics[width=1\linewidth]{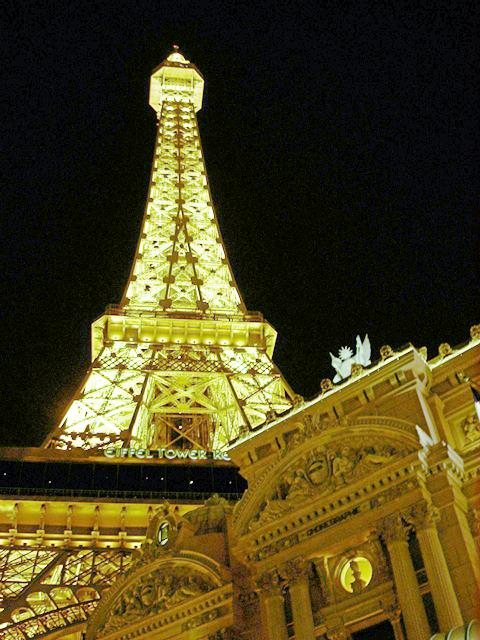}
        \caption*{}
    \end{subfigure} %

\vspace{-1.2em}
        \begin{subfigure}{.12\textwidth}
      \centering
        \includegraphics[width=1\linewidth]{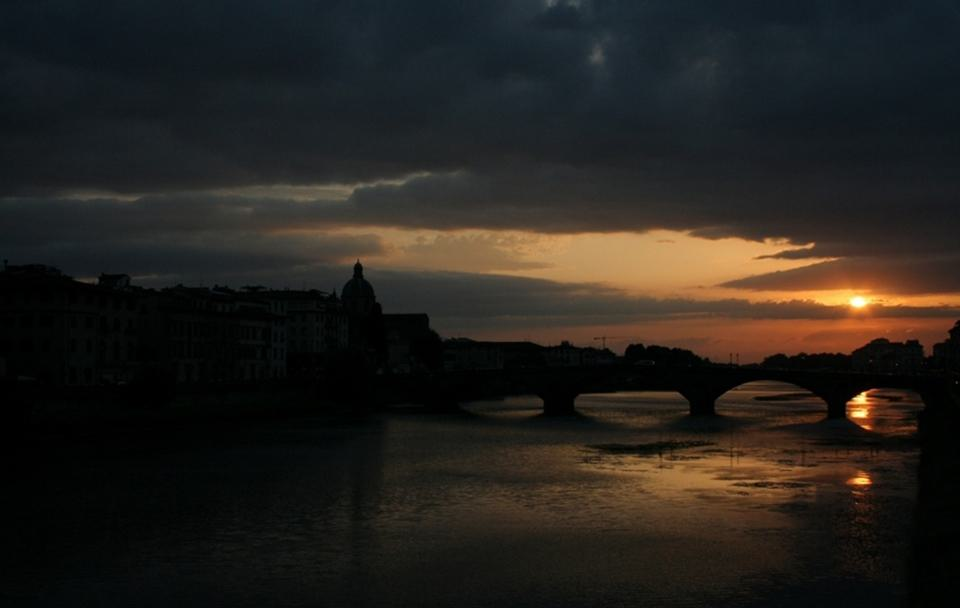}
        \caption*{}
    \end{subfigure} %
    \begin{subfigure}{.12\textwidth}
      \centering
        \includegraphics[width=1\linewidth]{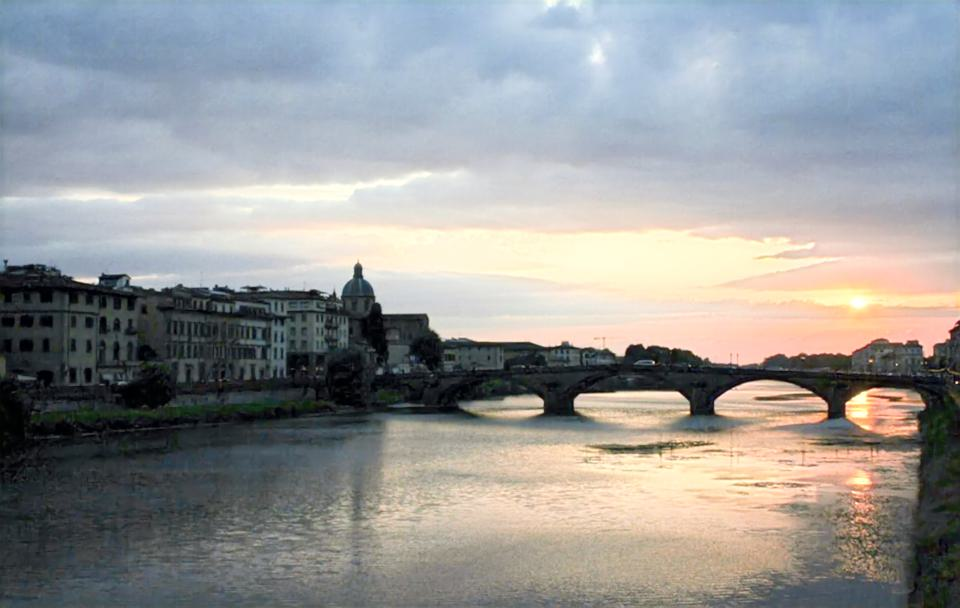}
        \caption*{}
    \end{subfigure} %
    \begin{subfigure}{.12\textwidth}
      \centering
        \includegraphics[width=1\linewidth]{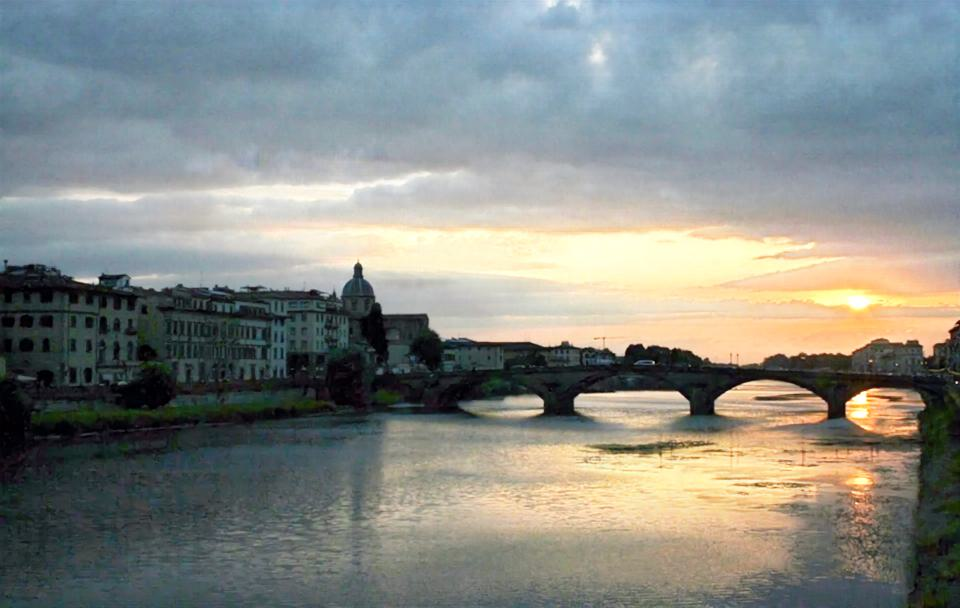}
        \caption*{}
    \end{subfigure} %
    \begin{subfigure}{.12\textwidth}
      \centering
        \includegraphics[width=1\linewidth]{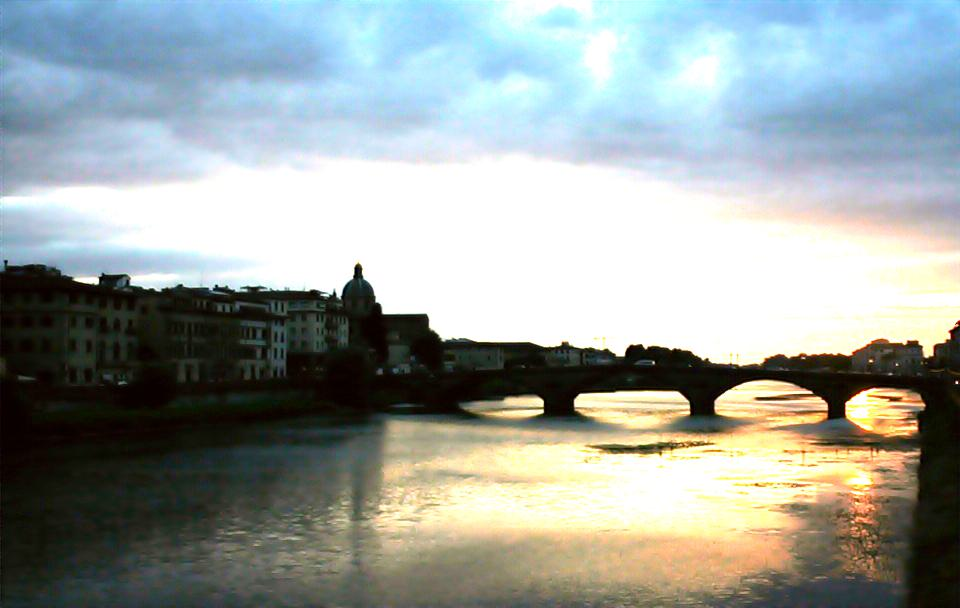}
        \caption*{}
    \end{subfigure} %
    \begin{subfigure}{.12\textwidth}
      \centering
        \includegraphics[width=1\linewidth]{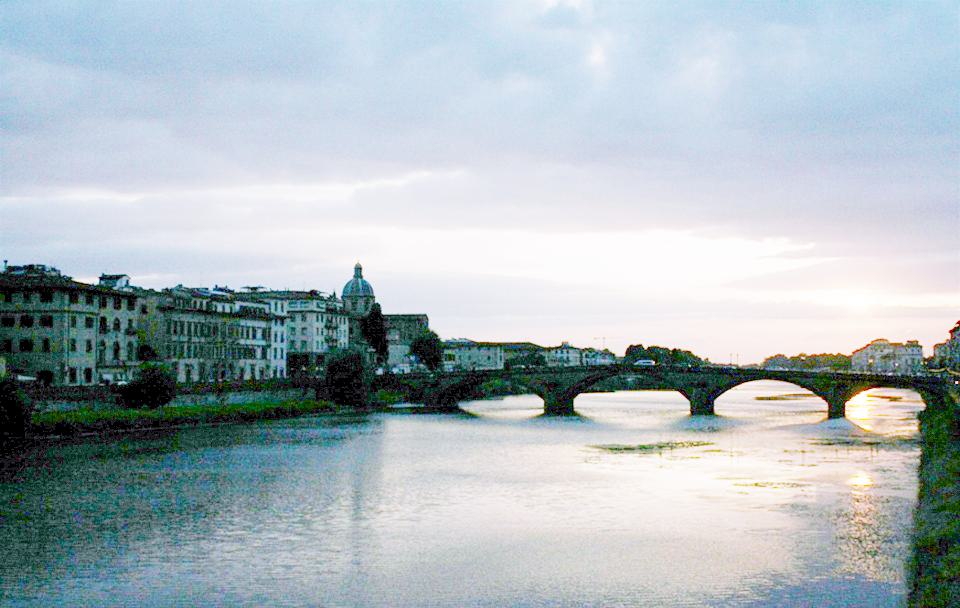}
        \caption*{}
    \end{subfigure} %
    \begin{subfigure}{.12\textwidth}
      \centering
        \includegraphics[width=1\linewidth]{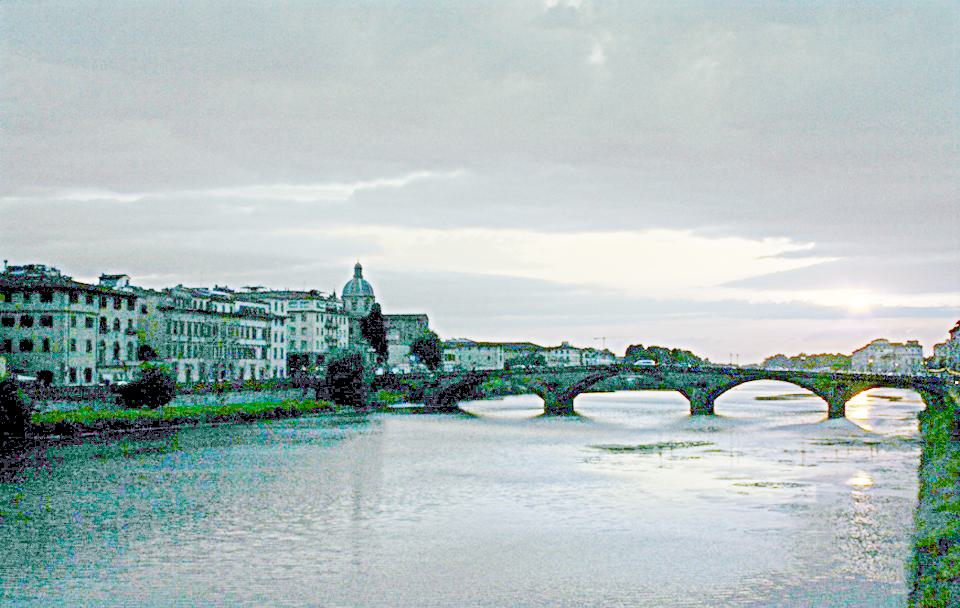}
        \caption*{}
    \end{subfigure} %
    \begin{subfigure}{.12\textwidth}
      \centering
        \includegraphics[width=1\linewidth]{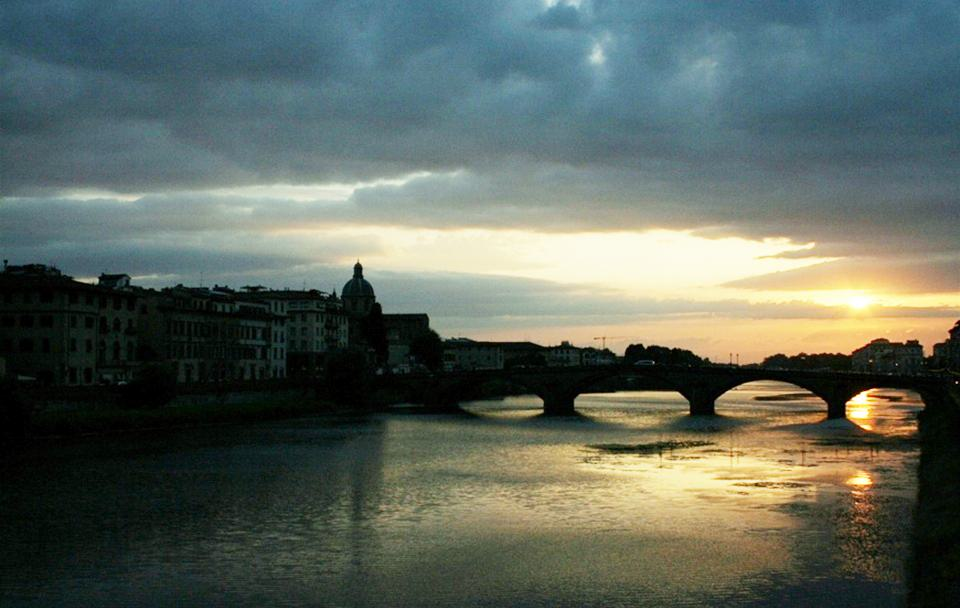}
        \caption*{}
    \end{subfigure} %
    \begin{subfigure}{.12\textwidth}
      \centering
        \includegraphics[width=1\linewidth]{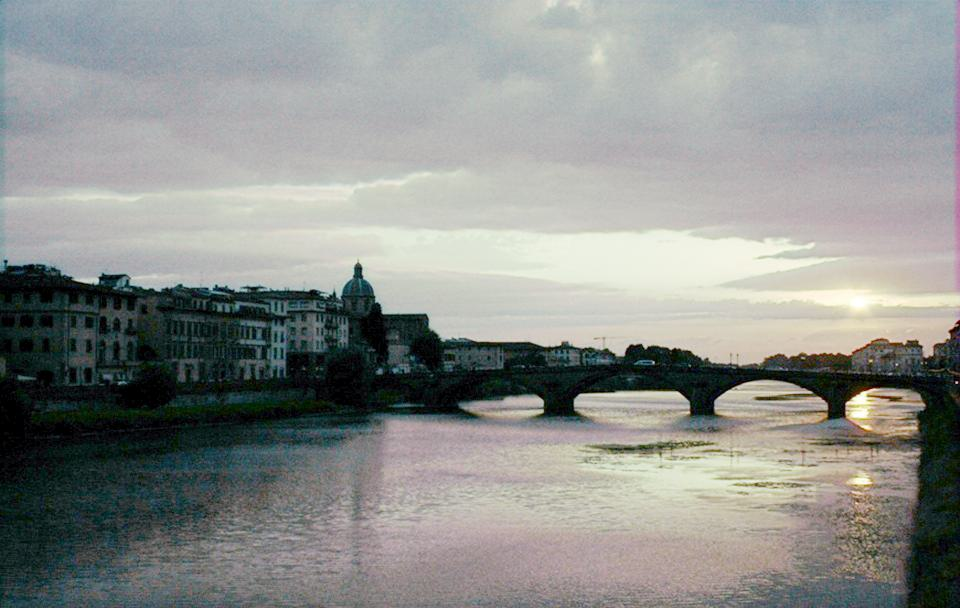}
        \caption*{}
    \end{subfigure} %

\vspace{-1.2em}

    \begin{subfigure}{.12\textwidth}
      \centering
        \caption*{Input}
    \end{subfigure} %
    \begin{subfigure}{.12\textwidth}
      \centering
        \caption*{ \tiny DRBN\cite{yang2020fidelity}\\+SKF\cite{Wu_2023_CVPR} }
    \end{subfigure} %
    \begin{subfigure}{.12\textwidth}
      \centering
        \caption*{\tiny SNR\cite{SNRAware}+\\SKF\cite{Wu_2023_CVPR} }
    \end{subfigure} %
    \begin{subfigure}{.12\textwidth}
      \centering
        \caption*{RUAS \cite{liu2021retinex} }
    \end{subfigure} %
    \begin{subfigure}{.12\textwidth}
      \centering
        \caption*{SGZ \cite{zheng2022semantic}}
    \end{subfigure} %
    \begin{subfigure}{.12\textwidth}
      \centering
        \caption*{Zero-DCE \cite{guo2020zero} }
    \end{subfigure} %
    \begin{subfigure}{.12\textwidth}
      \centering
        \caption*{SCI \cite{ma2022toward}}
    \end{subfigure} %
    \begin{subfigure}{.12\textwidth}
      \centering
        \caption*{Ours}
    \end{subfigure} %

    \caption{Qualitative comparison with related methods, all trained using the LOL \cite{wei2018deep}} dataset.
    \label{fig:comparison-qualitative}
\end{figure*}

\begin{table*}[]
\centering
\scriptsize

\begin{tabular}{cc|cccccc|cc}
\hline

Method & Requires & NOD \cite{morawski2021nod} & NOD SE \cite{morawski2021nod} & LOD \cite{hong2021crafting} & ExDark \cite{loh2019getting} & ExDark \cite{loh2019getting} & DarkFace \cite{poor_visibility_benchmark} & FLOPS &  \\
 & paired data & mAP $\uparrow$ & mAP $\uparrow$ & mAP $\uparrow$ & mAP $\uparrow$ & class. acc. $\uparrow$ & mAP@.5 $\uparrow$ & MACs $\downarrow$ & Params. $\downarrow$ \\ \hline

\rule{0pt}{2ex} (no enhancement) & N/A & 42.1\% & 22.8\% & 37.8\% & \firstbest{40.3\%} & 22.8\% & 33.3\% & - & - \\

Histogram Equalization & \checkmarkx & 39.8\% & 18.4\% & 38.9\% & 33.3\% & 39.9\% & 32.9\% & - & - \\
 % NL Means Denoising + Hist. Eq. & \checkmarkx & 36.6\% & 31.7\% & 33.0\% & 37.9\% & 27.5\% & 0 & 0 \\

 \hline

\rule{0pt}{3ex} DRBN\cite{yang2020fidelity}+SKF\cite{Wu_2023_CVPR} \textsuperscript{(CVPR20+CVPR23)} & \checkmark & \firstbest{43.1\%} & \secondbest{25.1\%} &  {46.5\%} & \secondbest{39.5\%} & 38.5\% & \firstbest{40.7\%} & 560G & 887M \\ %\textsubscript{CVPR20+CVPR23}

SNR\cite{SNRAware}+SKF\cite{Wu_2023_CVPR} \textsuperscript{(CVPR22+CVPR23} & \checkmark & 41.6\% & 20.1\% &  41.8\% & 39.0\% & 39.6\% & 37.2\% & 887G & 105M \\ %\textsubscript{CVPR22+CVPR23}

\hline

\rule{0pt}{3ex} RUAS \cite{liu2021retinex} \textsuperscript{(CVPR21)}& \checkmarkx & 38.6\% & 21.3\% & 41.1\% & 36.7\% & 34.1\% & 21.6\% & 4G & 3.4K\\

SGZ \cite{zheng2022semantic} \textsuperscript{(WACV22)}& \checkmarkx & 41.3\% & 23.9\% & 42.6\% & 34.7\% & 41.7\% & 37.0\% & 11G & 11K \\ %\textsubscript{WACV22} 

Zero-DCE \cite{guo2020zero} \textsuperscript{(CVPR20)} & \checkmarkx & 40.4\% & 20.5\% & 38.8\% & 32.8\% & {43.1\%} & 33.3\% & 83G & 79K \\ %\textsubscript{CVPR20} 

SCI \cite{ma2022toward} \textsuperscript{(CVPR22)} & \checkmarkx & \secondbest{42.9\%} & \firstbest{26.1\%} & \secondbest{46.9\%} & 38.7\% & \secondbest{43.8\%} & \secondbest{39.9\%} & 0.37G & 0.36K \\ %\textsubscript{CVPR20} 

Ours & \checkmarkx & {42.8\%} & \firstbest{26.1\%} & \firstbest{47.0\%} & 38.5\% & \firstbest{47.86\%} & {37.3\%} & 83G & 79K \\

\hline
\end{tabular}
\\ \colorlegend

\caption{Qualitative comparisons of the proposed and related works in terms of task-based performance. All methods were trained on the LOL \cite{wei2018deep} dataset that comprises 500 paired low- and normal-light images.}
\label{tab:comparison}
\end{table*}

\textbf{Semantic guidance}. We propose to leverage the zero-shot capabilities of the CLIP model to introduce semantic guidance during training in a straightforward way. Thanks to the simplicity, the proposed semantic guidance method can be used with any low-light dataset containing bounding-box annotations or any type of annotation that can be used to extract patches with an object category, which is advantageous given the sparsity and high collection costs of low-light datasets. During the training, we use annotation to extract image patches counting objects of a class ${cls}$. Because we use CLIP to perform zero-shot classification, the label set is not pre-fixed or limited and can be adjusted for a single batch dynamically during the training. For each instance, we use a pair of positive and negative class prompts, $T=\{$\textit{"a photo of a \(cls\)","not a photo of a \(cls\)"\}}$\}$ and experimentally show that this results in improved task-based performance of our method.

Given an enhanced low-light image $\hat{I}$ and its class label ${cls}$, we perform the classification based on the cosine similarity between the embedded enhanced image $\Phi_{img}(\hat{I})$ and a pair of antonym prompts $\Phi_{txt}(T)$, where $T=\{$\textit{"a photo of a \(cls\)","not a photo of a \(cls\)"\}}$\}$ is a pair of positive and negative prompts. 

\begin{equation}
  \hat{y}_{cls} = \frac
  {e^{cos(\Phi_{img}(\hat{I}), \Phi_{txt}(P_{cls})}}
  {\sum_{i \in \{cls, \neg cls\}} 
 e^{cos(\Phi_{img}(\hat{I}), \Phi_{txt}(P_i)}},
  \label{eq:prompt-image-cs-2}
\end{equation}

where $P_{cls}$ and $P_{\neg cls}$ is an antonym pair of positive and negative class prompts, "a photo of a \(cls\)" and "not a photo of a \(cls\)".

We use binary cross-entropy loss to guide the enhancement process in a way that facilitates high-level machine cognition. 

In contrast with methods that perform semantic guidance using pre-trained normal-light high-level models, our method can be applied to datasets without paired normal-light ground-truth data. Moreover, our method leverages the zero-shot capabilities of CLIP and is scalable to any dataset with at least bounding box-level annotation via open-vocabulary classification. Additionally, the method is easily extendable to unannotated paired low- and normal-light datasets by extracting high-confidence detections using a normal-light object detector.

\textbf{Total loss} The total training loss is expressed as:

\begin{equation}
\begin{split}
  L_{total} = 
  \lambda_{exp} L_{exp} + 
  \lambda_{spa} L_{spa} + \\
  \lambda_{RGB} L_{RGB} + 
  \lambda_{TV_{A}} L_{TV_{A}} + \\
  \lambda_{cls} L_{cls} + \lambda_{prompt} L_{prompt},
  \label{eq:loss-total}
\end{split}
\end{equation}

where $L_{exp}$, $L_{spa}$, $L_{RGB}$, $L{TV_{A}}$ are the zero-reference losses described in subsection \ref{sssec:zero-reference-losses}, $L_{cls}$ and $L_{prompt}$ are the proposed losses described in subsections \ref{sssec:leveraging-clip-training}, and $\lambda$ are the corresponding loss weights.
%-------------------------------------------------------------------------

\section{Experiments}
\label{sec:experimental_results}

In our work, we focus on improving low-light cognition by enhancing images before processing with downstream models. Similar to related work \cite{morawski2021nod,morawski2022genisp,wang2023tienet,hashmi2023featenhancer,robidoux2021end,yoshimura2023dynamicisp}, rather than assuming correlation between human perception and downstream task performances, we focus on task-based evaluation. Such evaluation has been used previously in \cite{morawski2021nod,morawski2022genisp,wang2023tienet,hashmi2023featenhancer}.

\label{ssec:experimental_results_implementation_details}
\textbf{Implementation details}\footnote{The experiments were completed by authors at National Taiwan University.}. We use Zero-DCE \cite{guo2020zero} as our baseline model and  CLIP for image classification and the learned prompt. For training the prompt, we learn the positive and negative prompts of length 16 using brightness, contrast and hue augmentation, and down-sample image crops by the scale factor of $4$ for positive samples and down-sample by nearest-neighbor downsampling by the factor of $4$ for negative samples. For training the enhancement model, we extract image patches containing objects based on dataset bounding box annotation and include a portion of the image around the bounding box if the object is very small (e.g. in the DARKFACE \cite{poor_visibility_benchmark} dataset). For paired low- and normal-light datasets without annotated object instances, we ran an open-vocabulary detector YOLO-World \cite{cheng2024yolo} on paired normal-light data using 365 labels in the Objects365 \cite{shao2019objects365} dataset and extract patches based on predictions with a confidence score over $0.3$. We train our enhancement network on $224 \times 224$ image patches with a batch size $8$, for $105K$ steps, using the Adam optimizer with a learning rate $0.0001$, weight decay $0.0001$ and gradient norm clipping set to $0.1$. We use zero-reference loss terms proposed by Guo \textit{et al.} \cite{Guo_2020_CVPR} and set the the weights of color loss term to $5$, spatial consistency term to $1$, exposure control term to $10$, and TV term $200$.

\subsection{Ablation Study} 
\label{ssec:experimental_results_ablation_study}
We conduct an extensive ablation study on multiple low-light datasets to show the contribution of each of the proposed loss terms. Following \cite{guo2020zero} recommendations, we train the model using a variety of different lighting condition. To this end, we use a collection of data extracted from the NOD \cite{morawski2021nod}, ExDark \cite{loh2019getting}, DarkFace \cite{poor_visibility_benchmark}, ExLPose \cite{ExLPose_2023_CVPR}, LOL \cite{wei2018deep} and BAID \cite{lv2022backlitnet} datasets to train all of the variants in Tab. \ref{tab:ablation}. To further increase the diversity of data, we use paired low- and normal-light datasets by using an open-vocabulary detector on normal-light data. For the LOL \cite{wei2018deep} and BAID \cite{lv2022backlitnet} datasets which do not contain any instance annotation, we extract object instances from low-light images based on paired detections on normal-light data with a confidence score over $0.3$ using YOLO-World \cite{cheng2024yolo}, using 365 labels in the Objects365 \cite{shao2019objects365} dataset. In addition to a large variety of illumination condition, our collection has a variety of different labels categories, and our method leverages zero-shot capabilities of the CLIP \cite{radford2021learning} model by adjusting the class prompts for each batch on-line. We show the basics dataset statistics in Fig. \ref{fig:training-stats}.

\textbf{Task-based performance}. As seen in the ablation study results presented in Tab. \ref{tab:ablation}, the proposed open-vocabulary classification loss leads to consistent improvements over the baseline method \cite{guo2020zero}, except for the DarkFace \cite{poor_visibility_benchmark} dataset. Similarly, introducing the proposed learned prompt leads to consistent improvement over both the baseline method \cite{guo2020zero} and detection with unenhanced images. The two proposed improvements together lead to consistent improvements over the baseline method \cite{guo2020zero} except the ExDark dataset \cite{loh2019getting} where the differences between the methods in the study are relatively small and well above detection performance using unenhanced images. Possible explanations for the observed difference for the ExDark \cite{loh2019getting} dataset include small proportion of the ExDark data during training or relatively higher average brightness levels in the dataset as seen in Fig. \ref{fig:training-stats}.

\textbf{Qualitative results.} Qualitative results of the ablation study are shown in Fig. \ref{fig:ablation-qualitative}. Semantic guidance improves the color distribution of the images, removing the color hue shift visible in images produced by the baseline method. The learned prompt improves the color distribution of the images and contributes to improving the contrast in the images. Compared with the baseline method, the learned prompt reduces the overexposure that leads to information loss from the original input image. Moreover, the learned prompt reduces over-amplification of the noise in the images.

\subsection{Task-based Comparison with Related Methods } 
\label{ssec:comparison}
We show task-based performance comparison with related methods in Tab. \ref{tab:comparison}, all trained on the LOL \cite{wei2018deep}.  We first compare our supervised methods DRBN \cite{yang2020fidelity} and SNR-LLNet \cite{SNRAware}
 with semantic guidance \cite{Wu_2023_CVPR}. Our method is on-par with DRBN+SKF\cite{yang2020fidelity,Wu_2023_CVPR} and outperforms SNR-LLNet \cite{SNRAware} in terms of task-based performance across the tested datasets. However, if compared to the computational complexity of the DRBN+SKF\cite{yang2020fidelity,Wu_2023_CVPR} our method relies on a much more lightweight architecture \cite{guo2020zero}. Similarly to the ablation study, our method shows improvement in term of task-based performance across all tested datasets except the ExDark \cite{loh2019getting} dataset. Again, ossible explanations for the observed difference for the ExDark \cite{loh2019getting} dataset include small proportion of the ExDark data during training or relatively higher average brightness levels in the dataset as seen in Fig. \ref{fig:training-stats}.

%-------------------------------------------------------------------------
\section{Conclusion}
\label{sec:conclusion}
We proposed to leverage the rich CLIP prior and CLIP's zero-shot capabilities at the training stage to improve zero-reference low-light image enhancement. We proposed to first pre-train a pair of prompts that capture enhanced low-light image prior via prompt learning with a simple data augmentation strategy without any need for paired or unpaired normal-light data. We experimentally showed that the learned prompt helps guiding the enhancement by improving the image contrast, reducing over-enhancement and reducing over-amplification of noise. Next, we proposed to further reuse the CLIP model during the training process using a straightforward yet effective and scalable semantic segmentation via zero-shot open vocabulary classification. We conducted extensive experiments that show consistent improvements over the baseline method across various low-light datasets in terms of task-based performance.
%-------------------------------------------------------------------------
\section*{Acknowledgements}
\label{sec:acknowledgements}
This work was supported in part by National Science and Technology Council, Taiwan, under Grant NSTC 111-2634-F-002-022 and by Qualcomm through a Taiwan University Research Collaboration Project.

%%%%%%%%% REFERENCES
{\small
\bibliographystyle{ieee_fullname}
\bibliography{Paper}

\begin{thebibliography}{10}\itemsep=-1pt

\bibitem{aakerberg2022semantic}
Andreas Aakerberg, Anders~S Johansen, Kamal Nasrollahi, and Thomas~B Moeslund.
\newblock Semantic segmentation guided real-world super-resolution.
\newblock In {\em Proceedings of the IEEE/CVF Winter Conference on Applications of Computer Vision}, pages 449--458, 2022.

\bibitem{buchsbaum1980spatial}
Gershon Buchsbaum.
\newblock A spatial processor model for object colour perception.
\newblock {\em Journal of the Franklin institute}, 310(1):1--26, 1980.

\bibitem{bychkovsky2011learning}
Vladimir Bychkovsky, Sylvain Paris, Eric Chan, and Fr{\'e}do Durand.
\newblock Learning photographic global tonal adjustment with a database of input/output image pairs.
\newblock In {\em CVPR 2011}, pages 97--104. IEEE, 2011.

\bibitem{cai2018learning}
Jianrui Cai, Shuhang Gu, and Lei Zhang.
\newblock Learning a deep single image contrast enhancer from multi-exposure images.
\newblock {\em IEEE Transactions on Image Processing}, 27(4):2049--2062, 2018.

\bibitem{chen2019seeing}
Chen Chen, Qifeng Chen, Minh~N Do, and Vladlen Koltun.
\newblock Seeing motion in the dark.
\newblock In {\em Proceedings of the IEEE/CVF International conference on computer vision}, pages 3185--3194, 2019.

\bibitem{chen2018learning}
Chen Chen, Qifeng Chen, Jia Xu, and Vladlen Koltun.
\newblock Learning to see in the dark.
\newblock In {\em Proceedings of the IEEE conference on computer vision and pattern recognition}, pages 3291--3300, 2018.

\bibitem{cheng2024yolo}
Tianheng Cheng, Lin Song, Yixiao Ge, Wenyu Liu, Xinggang Wang, and Ying Shan.
\newblock Yolo-world: Real-time open-vocabulary object detection.
\newblock {\em arXiv preprint arXiv:2401.17270}, 2024.

\bibitem{dong2022abandoning}
Xingbo Dong, Wanyan Xu, Zhihui Miao, Lan Ma, Chao Zhang, Jiewen Yang, Zhe Jin, Andrew Beng~Jin Teoh, and Jiajun Shen.
\newblock Abandoning the bayer-filter to see in the dark.
\newblock In {\em Proceedings of the IEEE/CVF Conference on Computer Vision and Pattern Recognition}, pages 17431--17440, 2022.

\bibitem{fan2022half}
Chi-Mao Fan, Tsung-Jung Liu, and Kuan-Hsien Liu.
\newblock Half wavelet attention on m-net+ for low-light image enhancement.
\newblock In {\em 2022 IEEE International Conference on Image Processing (ICIP)}, pages 3878--3882. IEEE, 2022.

\bibitem{fu2023you}
Huiyuan Fu, Wenkai Zheng, Xiangyu Meng, Xin Wang, Chuanming Wang, and Huadong Ma.
\newblock You do not need additional priors or regularizers in retinex-based low-light image enhancement.
\newblock In {\em Proceedings of the IEEE/CVF Conference on Computer Vision and Pattern Recognition}, pages 18125--18134, 2023.

\bibitem{fu2016fusion}
Xueyang Fu, Delu Zeng, Yue Huang, Yinghao Liao, Xinghao Ding, and John Paisley.
\newblock A fusion-based enhancing method for weakly illuminated images.
\newblock {\em Signal Processing}, 129:82--96, 2016.

\bibitem{gonzalez2009digital}
Rafael~C Gonzalez.
\newblock {\em Digital image processing}.
\newblock Pearson education india, 2009.

\bibitem{guo2020zero}
Chunle Guo, Chongyi Li, Jichang Guo, Chen~Change Loy, Junhui Hou, Sam Kwong, and Runmin Cong.
\newblock Zero-reference deep curve estimation for low-light image enhancement.
\newblock In {\em Proceedings of the IEEE/CVF conference on computer vision and pattern recognition}, pages 1780--1789, 2020.

\bibitem{Guo_2020_CVPR}
Chunle Guo, Chongyi Li, Jichang Guo, Chen~Change Loy, Junhui Hou, Sam Kwong, and Runmin Cong.
\newblock Zero-reference deep curve estimation for low-light image enhancement.
\newblock In {\em Proceedings of the IEEE/CVF Conference on Computer Vision and Pattern Recognition (CVPR)}, June 2020.

\bibitem{guo2016lime}
Xiaojie Guo, Yu Li, and Haibin Ling.
\newblock Lime: Low-light image enhancement via illumination map estimation.
\newblock {\em IEEE Transactions on image processing}, 26(2):982--993, 2016.

\bibitem{hai2023r2rnet}
Jiang Hai, Zhu Xuan, Ren Yang, Yutong Hao, Fengzhu Zou, Fang Lin, and Songchen Han.
\newblock R2rnet: Low-light image enhancement via real-low to real-normal network.
\newblock {\em Journal of Visual Communication and Image Representation}, 90:103712, 2023.

\bibitem{hashmi2023featenhancer}
Khurram~Azeem Hashmi, Goutham Kallempudi, Didier Stricker, and Muhammad~Zeshan Afzal.
\newblock Featenhancer: Enhancing hierarchical features for object detection and beyond under low-light vision.
\newblock In {\em Proceedings of the IEEE/CVF International Conference on Computer Vision}, pages 6725--6735, 2023.

\bibitem{hong2021crafting}
Yang Hong, Kaixuan Wei, Linwei Chen, and Ying Fu.
\newblock Crafting object detection in very low light.
\newblock In {\em BMVC}, volume~1, page~3, 2021.

\bibitem{jiang2019learning}
Haiyang Jiang and Yinqiang Zheng.
\newblock Learning to see moving objects in the dark.
\newblock In {\em Proceedings of the IEEE/CVF International Conference on Computer Vision}, pages 7324--7333, 2019.

\bibitem{jiang2021enlightengan}
Yifan Jiang, Xinyu Gong, Ding Liu, Yu Cheng, Chen Fang, Xiaohui Shen, Jianchao Yang, Pan Zhou, and Zhangyang Wang.
\newblock Enlightengan: Deep light enhancement without paired supervision.
\newblock {\em IEEE transactions on image processing}, 30:2340--2349, 2021.

\bibitem{jobson1997multiscale}
Daniel~J Jobson, Zia-ur Rahman, and Glenn~A Woodell.
\newblock A multiscale retinex for bridging the gap between color images and the human observation of scenes.
\newblock {\em IEEE Transactions on Image processing}, 6(7):965--976, 1997.

\bibitem{jobson1997properties}
Daniel~J Jobson, Zia-ur Rahman, and Glenn~A Woodell.
\newblock Properties and performance of a center/surround retinex.
\newblock {\em IEEE transactions on image processing}, 6(3):451--462, 1997.

\bibitem{kim2001advanced}
Joung-Youn Kim, Lee-Sup Kim, and Seung-Ho Hwang.
\newblock An advanced contrast enhancement using partially overlapped sub-block histogram equalization.
\newblock {\em IEEE transactions on circuits and systems for video technology}, 11(4):475--484, 2001.

\bibitem{kuo2022f}
Weicheng Kuo, Yin Cui, Xiuye Gu, AJ Piergiovanni, and Anelia Angelova.
\newblock F-vlm: Open-vocabulary object detection upon frozen vision and language models.
\newblock {\em arXiv preprint arXiv:2209.15639}, 2022.

\bibitem{land1977retinex}
Edwin~H Land.
\newblock The retinex theory of color vision.
\newblock {\em Scientific american}, 237(6):108--129, 1977.

\bibitem{ExLPose_2023_CVPR}
Sohyun Lee, Jaesung Rim, Boseung Jeong, Geonu Kim, ByungJu Woo, Haechan Lee, and Suha~Kwak Sunghyun~Cho.
\newblock Human pose estimation in extremely low-light conditions.
\newblock In {\em Proceedings of the IEEE/CVF Conference on Computer Vision and Pattern Recognition (CVPR)}, 2023.

\bibitem{li2021learning}
Chongyi Li, Chunle Guo, and Chen~Change Loy.
\newblock Learning to enhance low-light image via zero-reference deep curve estimation.
\newblock {\em IEEE Transactions on Pattern Analysis and Machine Intelligence}, 44(8):4225--4238, 2021.

\bibitem{li2017joint}
Mading Li, Jiaying Liu, Wenhan Yang, and Zongming Guo.
\newblock Joint denoising and enhancement for low-light images via retinex model.
\newblock In {\em International Forum on Digital TV and Wireless Multimedia Communications}, pages 91--99. Springer, 2017.

\bibitem{li2018structure}
Mading Li, Jiaying Liu, Wenhan Yang, Xiaoyan Sun, and Zongming Guo.
\newblock Structure-revealing low-light image enhancement via robust retinex model.
\newblock {\em IEEE Transactions on Image Processing}, 27(6):2828--2841, 2018.

\bibitem{li2020blind}
Xiaoming Li, Chaofeng Chen, Shangchen Zhou, Xianhui Lin, Wangmeng Zuo, and Lei Zhang.
\newblock Blind face restoration via deep multi-scale component dictionaries.
\newblock In {\em European conference on computer vision}, pages 399--415. Springer, 2020.

\bibitem{liang2021recurrent}
Jinxiu Liang, Jingwen Wang, Yuhui Quan, Tianyi Chen, Jiaying Liu, Haibin Ling, and Yong Xu.
\newblock Recurrent exposure generation for low-light face detection.
\newblock {\em IEEE Transactions on Multimedia}, 24:1609--1621, 2021.

\bibitem{liang2023iterative}
Zhexin Liang, Chongyi Li, Shangchen Zhou, Ruicheng Feng, and Chen~Change Loy.
\newblock Iterative prompt learning for unsupervised backlit image enhancement.
\newblock In {\em Proceedings of the IEEE/CVF International Conference on Computer Vision}, pages 8094--8103, 2023.

\bibitem{liu2017image}
Ding Liu, Bihan Wen, Xianming Liu, Zhangyang Wang, and Thomas~S Huang.
\newblock When image denoising meets high-level vision tasks: A deep learning approach.
\newblock {\em arXiv preprint arXiv:1706.04284}, 2017.

\bibitem{liu2021retinex}
Risheng Liu, Long Ma, Jiaao Zhang, Xin Fan, and Zhongxuan Luo.
\newblock Retinex-inspired unrolling with cooperative prior architecture search for low-light image enhancement.
\newblock In {\em Proceedings of the IEEE/CVF Conference on Computer Vision and Pattern Recognition}, pages 10561--10570, 2021.

\bibitem{loh2019getting}
Yuen~Peng Loh and Chee~Seng Chan.
\newblock Getting to know low-light images with the exclusively dark dataset.
\newblock {\em Computer Vision and Image Understanding}, 178:30--42, 2019.

\bibitem{lv2022backlitnet}
Xiaoqian Lv, Shengping Zhang, Qinglin Liu, Haozhe Xie, Bineng Zhong, and Huiyu Zhou.
\newblock Backlitnet: A dataset and network for backlit image enhancement.
\newblock {\em Computer Vision and Image Understanding}, 218:103403, 2022.

\bibitem{ma2022toward}
Long Ma, Tengyu Ma, Risheng Liu, Xin Fan, and Zhongxuan Luo.
\newblock Toward fast, flexible, and robust low-light image enhancement.
\newblock In {\em Proceedings of the IEEE/CVF Conference on Computer Vision and Pattern Recognition}, pages 5637--5646, 2022.

\bibitem{morawski2022genisp}
Igor Morawski, Yu-An Chen, Yu-Sheng Lin, Shusil Dangi, Kai He, and Winston~H Hsu.
\newblock Genisp: neural isp for low-light machine cognition.
\newblock In {\em Proceedings of the IEEE/CVF Conference on Computer Vision and Pattern Recognition}, pages 630--639, 2022.

\bibitem{morawski2021nod}
Igor Morawski, Yu-An Chen, Yu-Sheng Lin, and Winston~H Hsu.
\newblock Nod: Taking a closer look at detection under extreme low-light conditions with night object detection dataset.
\newblock {\em arXiv preprint arXiv:2110.10364}, 2021.

\bibitem{radford2021learning}
Alec Radford, Jong~Wook Kim, Chris Hallacy, Aditya Ramesh, Gabriel Goh, Sandhini Agarwal, Girish Sastry, Amanda Askell, Pamela Mishkin, Jack Clark, et~al.
\newblock Learning transferable visual models from natural language supervision.
\newblock In {\em International conference on machine learning}, pages 8748--8763. PMLR, 2021.

\bibitem{robidoux2021end}
Nicolas Robidoux, Luis E~Garcia Capel, Dong-eun Seo, Avinash Sharma, Federico Ariza, and Felix Heide.
\newblock End-to-end high dynamic range camera pipeline optimization.
\newblock In {\em Proceedings of the IEEE/CVF Conference on Computer Vision and Pattern Recognition}, pages 6297--6307, 2021.

\bibitem{shao2019objects365}
Shuai Shao, Zeming Li, Tianyuan Zhang, Chao Peng, Gang Yu, Xiangyu Zhang, Jing Li, and Jian Sun.
\newblock Objects365: A large-scale, high-quality dataset for object detection.
\newblock In {\em Proceedings of the IEEE/CVF international conference on computer vision}, pages 8430--8439, 2019.

\bibitem{stark2000adaptive}
J~Alex Stark.
\newblock Adaptive image contrast enhancement using generalizations of histogram equalization.
\newblock {\em IEEE Transactions on image processing}, 9(5):889--896, 2000.

\bibitem{wang2023exploring}
Jianyi Wang, Kelvin~CK Chan, and Chen~Change Loy.
\newblock Exploring clip for assessing the look and feel of images.
\newblock In {\em Proceedings of the AAAI Conference on Artificial Intelligence}, volume~37, pages 2555--2563, 2023.

\bibitem{wang2007fast}
Qing Wang and Rabab~K Ward.
\newblock Fast image/video contrast enhancement based on weighted thresholded histogram equalization.
\newblock {\em IEEE transactions on Consumer Electronics}, 53(2):757--764, 2007.

\bibitem{wang2013naturalness}
Shuhang Wang, Jin Zheng, Hai-Miao Hu, and Bo Li.
\newblock Naturalness preserved enhancement algorithm for non-uniform illumination images.
\newblock {\em IEEE transactions on image processing}, 22(9):3538--3548, 2013.

\bibitem{wang2021hla}
Wenjing Wang, Wenhan Yang, and Jiaying Liu.
\newblock Hla-face: Joint high-low adaptation for low light face detection.
\newblock In {\em Proceedings of the IEEE/CVF Conference on Computer Vision and Pattern Recognition}, pages 16195--16204, 2021.

\bibitem{wang2018recovering}
Xintao Wang, Ke Yu, Chao Dong, and Chen~Change Loy.
\newblock Recovering realistic texture in image super-resolution by deep spatial feature transform.
\newblock In {\em Proceedings of the IEEE conference on computer vision and pattern recognition}, pages 606--615, 2018.

\bibitem{wang2023tienet}
Yudong Wang, Jichang Guo, Ruining Wang, Wanru He, and Chongyi Li.
\newblock Tienet: task-oriented image enhancement network for degraded object detection.
\newblock {\em Signal, Image and Video Processing}, pages 1--8, 2023.

\bibitem{wang2022low}
Yufei Wang, Renjie Wan, Wenhan Yang, Haoliang Li, Lap-Pui Chau, and Alex Kot.
\newblock Low-light image enhancement with normalizing flow.
\newblock In {\em Proceedings of the AAAI conference on artificial intelligence}, volume~36, pages 2604--2612, 2022.

\bibitem{wei2018deep}
Chen Wei, Wenjing Wang, Wenhan Yang, and Jiaying Liu.
\newblock Deep retinex decomposition for low-light enhancement.
\newblock {\em arXiv preprint arXiv:1808.04560}, 2018.

\bibitem{wu2022uretinex}
Wenhui Wu, Jian Weng, Pingping Zhang, Xu Wang, Wenhan Yang, and Jianmin Jiang.
\newblock Uretinex-net: Retinex-based deep unfolding network for low-light image enhancement.
\newblock In {\em Proceedings of the IEEE/CVF conference on computer vision and pattern recognition}, pages 5901--5910, 2022.

\bibitem{Wu_2023_CVPR}
Yuhui Wu, Chen Pan, Guoqing Wang, Yang Yang, Jiwei Wei, Chongyi Li, and Heng~Tao Shen.
\newblock Learning semantic-aware knowledge guidance for low-light image enhancement.
\newblock In {\em Proceedings of the IEEE/CVF Conference on Computer Vision and Pattern Recognition (CVPR)}, pages 1662--1671, June 2023.

\bibitem{xu2020learning}
Ke Xu, Xin Yang, Baocai Yin, and Rynson~WH Lau.
\newblock Learning to restore low-light images via decomposition-and-enhancement.
\newblock In {\em Proceedings of the IEEE/CVF conference on computer vision and pattern recognition}, pages 2281--2290, 2020.

\bibitem{SNRAware}
Xiaogang Xu, Ruixing Wang, Chi-Wing Fu, and Jiaya Jia.
\newblock Snr-aware low-light image enhancement.
\newblock In {\em 2022 IEEE/CVF Conference on Computer Vision and Pattern Recognition (CVPR)}, pages 17693--17703, 2022.

\bibitem{Xu_2023_CVPR}
Xiaogang Xu, Ruixing Wang, and Jiangbo Lu.
\newblock Low-light image enhancement via structure modeling and guidance.
\newblock In {\em Proceedings of the IEEE/CVF Conference on Computer Vision and Pattern Recognition (CVPR)}, pages 9893--9903, June 2023.

\bibitem{xu2021arid}
Yuecong Xu, Jianfei Yang, Haozhi Cao, Kezhi Mao, Jianxiong Yin, and Simon See.
\newblock Arid: A new dataset for recognizing action in the dark.
\newblock In {\em Deep Learning for Human Activity Recognition: Second International Workshop, DL-HAR 2020, Held in Conjunction with IJCAI-PRICAI 2020, Kyoto, Japan, January 8, 2021, Proceedings 2}, pages 70--84. Springer, 2021.

\bibitem{yang2020fidelity}
Wenhan Yang, Shiqi Wang, Yuming Fang, Yue Wang, and Jiaying Liu.
\newblock From fidelity to perceptual quality: A semi-supervised approach for low-light image enhancement.
\newblock In {\em Proceedings of the IEEE/CVF conference on computer vision and pattern recognition}, pages 3063--3072, 2020.

\bibitem{yang2021sparse}
Wenhan Yang, Wenjing Wang, Haofeng Huang, Shiqi Wang, and Jiaying Liu.
\newblock Sparse gradient regularized deep retinex network for robust low-light image enhancement.
\newblock {\em IEEE Transactions on Image Processing}, 30:2072--2086, 2021.

\bibitem{poor_visibility_benchmark}
Wenhan Yang, Ye Yuan, Wenqi Ren, Jiaying Liu, Walter~J. Scheirer, Zhangyang Wang, Zhang, and et al.
\newblock Advancing image understanding in poor visibility environments: A collective benchmark study.
\newblock {\em IEEE Transactions on Image Processing}, 29:5737--5752, 2020.

\bibitem{yoshimura2023dynamicisp}
Masakazu Yoshimura, Junji Otsuka, Atsushi Irie, and Takeshi Ohashi.
\newblock Dynamicisp: dynamically controlled image signal processor for image recognition.
\newblock In {\em Proceedings of the IEEE/CVF International Conference on Computer Vision}, pages 12866--12876, 2023.

\bibitem{zang2022open}
Yuhang Zang, Wei Li, Kaiyang Zhou, Chen Huang, and Chen~Change Loy.
\newblock Open-vocabulary detr with conditional matching.
\newblock In {\em European Conference on Computer Vision}, pages 106--122. Springer, 2022.

\bibitem{zhang2021beyond}
Yonghua Zhang, Xiaojie Guo, Jiayi Ma, Wei Liu, and Jiawan Zhang.
\newblock Beyond brightening low-light images.
\newblock {\em International Journal of Computer Vision}, 129:1013--1037, 2021.

\bibitem{zhang2019kindling}
Yonghua Zhang, Jiawan Zhang, and Xiaojie Guo.
\newblock Kindling the darkness: A practical low-light image enhancer.
\newblock In {\em Proceedings of the 27th ACM international conference on multimedia}, pages 1632--1640, 2019.

\bibitem{zhang2022deep}
Zhao Zhang, Huan Zheng, Richang Hong, Mingliang Xu, Shuicheng Yan, and Meng Wang.
\newblock Deep color consistent network for low-light image enhancement.
\newblock In {\em Proceedings of the IEEE/CVF conference on computer vision and pattern recognition}, pages 1899--1908, 2022.

\bibitem{zheng2021adaptive}
Chuanjun Zheng, Daming Shi, and Wentian Shi.
\newblock Adaptive unfolding total variation network for low-light image enhancement.
\newblock In {\em Proceedings of the IEEE/CVF international conference on computer vision}, pages 4439--4448, 2021.

\bibitem{zheng2022semantic}
Shen Zheng and Gaurav Gupta.
\newblock Semantic-guided zero-shot learning for low-light image/video enhancement.
\newblock In {\em Proceedings of the IEEE/CVF Winter conference on applications of computer vision}, pages 581--590, 2022.

\bibitem{zhou2022extract}
Chong Zhou, Chen~Change Loy, and Bo Dai.
\newblock Extract free dense labels from clip.
\newblock In {\em European Conference on Computer Vision}, pages 696--712. Springer, 2022.

\end{thebibliography}
}

\end{document}